%% file: main.tex
\documentclass[conference]{IEEEtran}
\IEEEoverridecommandlockouts

\usepackage{cite}
\usepackage{amsmath,amssymb,amsfonts}
\usepackage{algorithmic}
\usepackage{graphicx}
\usepackage{enumitem}
\usepackage{booktabs}
\usepackage{tabularx}
\usepackage{textcomp}
\usepackage{url}
\usepackage{xcolor}

\usepackage{makecell}
\usepackage{subcaption}
\usepackage[ruled,vlined]{algorithm2e}
\usepackage{hyperref}

\usepackage{booktabs}
\usepackage{multirow}
\definecolor{darkblue}{rgb}{0, 0, 0.5}
\hypersetup{colorlinks=true,citecolor=darkblue, linkcolor=darkblue, urlcolor=darkblue}
\usepackage{xurl} 

\input{frames}

\newcommand{\indsize}{\scriptsize}
\newcommand{\colind}[2]{\displaystyle\smash{\mathop{#1}^{\raisebox{.5\normalbaselineskip}{\indsize #2}}}}
\newcommand{\rowind}[1]{\mbox{\indsize #1}}

\usepackage{eso-pic}
\AddToShipoutPictureBG*{%
  \AtPageUpperLeft{\put(0,-20){\makebox[\paperwidth]{%
    \parbox{0.85\paperwidth}{\centering\scriptsize
    \copyright~2026 IEEE. Personal use of this material is permitted.
    Permission from IEEE must be obtained for all other uses, in any current
    or future media, including reprinting/republishing this material for
    advertising or promotional purposes, creating new collective works, for
    resale or redistribution to servers or lists, or reuse of any copyrighted
    component of this work in other works.
    Accepted at the 13th IEEE Special Session Privacy and Security of Big Data of IEEE BigData 2026.}}}}}

\begin{document}

\title{Pairing AIS with RF: Analysis of a Markovian Approach for Maritime Data Fusion
}

\author{
\IEEEauthorblockN{
Daniel Grimm\IEEEauthorrefmark{1},
Kyra Jongman\IEEEauthorrefmark{2},
Renato Cordeiro Ferreira\IEEEauthorrefmark{1},
Giuseppe Cascavilla\IEEEauthorrefmark{1},\\
Alfredo Cuzzocrea\IEEEauthorrefmark{3},
Willem-Jan Van Den Heuvel\IEEEauthorrefmark{1}
}
\IEEEauthorblockA{
\IEEEauthorrefmark{1}\IEEEauthorrefmark{2}Department of Intelligent Systems,
Tilburg University, Tilburg, The Netherlands\\
\IEEEauthorrefmark{3}University of Calabria, Rende, Italy\\
\IEEEauthorrefmark{1}\texttt{\{D.M.Grimm,R.CordeiroFerreira,g.cascavilla,W.J.A.M.vdnHeuvel\}@tilburguniversity.edu}\\
\IEEEauthorrefmark{3}\texttt{alfredo.cuzzocrea@unical.it}
}
}

\maketitle

\begin{abstract}
Monitoring maritime activities is important for safety and security, particularly for detecting illegal activities, but remains challenging due to the size and complexity of the maritime domain. The Automatic Identification System (AIS) is commonly used for this purpose, despite its known vulnerabilities. In contrast, space-based radio frequency (RF) observations provide an independent source of vessel activity but lack ship identification and therefore cannot identify vessels on their own. Following the CRISP-DM process, this study applies a Pair Hidden Markov Model (PHMM) to align individual RF detections with AIS trajectories, using temporal and spatial information while modeling noise and timing discrepancies. We evaluate it on NOAA AIS data with synthetic RF detections against nearest-neighbor baselines and analyze its errors with bias tests on trajectory length, sampling density, stationarity and AIS--RF distance. The length-corrected Forward score reaches a top-1 precision of $0.7776$ on the validation split of multi-candidate detections, where errors are possible, and an AUC of $0.637$ for detecting dark vessels among the detections that pass the prefilter. With a single RF detection, the score mostly reflects the AIS trajectory itself, and its errors favor longer and more stationary trajectories. A log-odds score, which compares the full model with a null model without match transitions, raises precision to $0.8580$ and AUC to $0.852$. Its length-corrected variant no longer shows a significant bias towards stationary trajectories and short inter-message distances, but still favors candidates farther from the RF detection. A time-weighted nearest-neighbor baseline without fitting remains ahead (precision $0.9013$, AUC $0.904$), so this single-detection, hand-designed PHMM formulation approaches but does not beat it.
\end{abstract}


\section{Introduction}
Global trade has relied on maritime transport for centuries. Today, an estimated $90\%$ of all international goods, ranging from raw materials to everyday products such as fruit and clothing, are transported by sea~\cite{pirozhnikov-2023}. Maritime transport is attractive because vessels can carry large volumes of cargo over long distances at relatively low cost. However, the growing reliance on this mode of transportation, combined with other maritime traffic, has led to increased congestion at sea~\cite{mou-2010}. This is particularly evident near ports and in narrow waterways, where dense traffic heightens the risk of accidents. In addition to congestion, the vast and largely unregulated nature of the oceans facilitates illicit activities such as smuggling, unreported fishing, and sanctions evasion. 

Although initially designed as a collision-avoidance tool, authorities rely heavily on the Automatic Identification System~(AIS)~\cite{meyer-2025,international-maritime-organization-imo-2003} for continuous vessel tracking. AIS transponders broadcast information such as a vessel’s Maritime Mobile Service Identity (MMSI) number, type, position, and course to nearby ships and coastal authorities. Despite its essential role, AIS is widely recognized to have significant limitations, including incomplete coverage and vulnerabilities to deliberate manipulation~\cite{balduzzi-2021}. Spoofing AIS messages to falsify identity or location, as well as intentionally deactivating transmissions, referred to as ``dark vessels", are well-known tactics of manipulation. For example, in July 2023, a Malaysian-flagged oil tanker was found to have spoofed its AIS signals to mask its true location, a tactic used for loading and transferring Iranian oil in violation of US sanctions~\cite{lloyds-2026}. Similarly, in late 2024, the Finnish Coast Guard reported that tankers had spoofed their location data to conceal visits to Russian ports~\cite{kauranen-2024}. 

As such, recent technical advances have introduced space-based radio frequency~(RF) detections as a promising complementary data source for vessel tracking. As integrating AIS and space-based RF data appears highly valuable, the use of statistical techniques for their data fusion should be explored. Therefore, the main research question behind this study is: \textit{How can AIS and RF data be matched to enhance ship identification?}

To address this question, this research proposes using a Pair Hidden Markov Model~(PHMM). The goal is to evaluate the potential of this statistical model for linking AIS data with space-based RF detections to improve vessel identification. This prompted the following two sub-questions:

\begin{Framed}
\textbf{RQ1:} How can a PHMM be designed to match AIS and RF data?
\end{Framed}

\begin{Framed}
\textbf{RQ2:} How does a PHMM compare with a tuning-free nearest-neighbor baseline for matching AIS and RF data on synthetic RF data, both in top-1 precision and in detecting dark vessels?
\end{Framed}

\pagebreak

To address the research questions, this study adopts the Cross-Industry Standard Process for Data Mining~(CRISP-DM) as a systematic methodology~\cite{Shimaoka:2024:IPT:CRISPDM-SLR}. The framework comprises six phases: (1) Business Understanding, (2) Data Understanding, (3) Data Preparation, (4) Modeling, (5) Evaluation, and (6) Deployment.

The sub-questions are mapped to the relevant phases of the CRISP-DM framework, which form the basis for the study’s structure. These phases are not strictly sequential, allowing revisions to earlier steps. The \textit{Deployment} phase is not included, as it falls outside the scope of this research. 

As a result, we contribute
(i) a PHMM for matching a single RF detection to AIS trajectory candidates, scored by the raw Forward score and by a log-odds score against a null model,
(ii) a comparison with nearest-neighbor baselines, in top-1 precision and in detecting dark vessels, on NOAA AIS data with synthetic RF detections, and
(iii) a bias analysis of the matching errors.
The hand-designed, single-detection formulation approaches but does not beat a
tuning-free time-weighted nearest-neighbor baseline.
All code developed for this study is available as a reproduction package on GitHub%
\footnote{\url{https://github.com/grimmmbo/AIS_RF_matching}}. 

\section{Background}
\subsection{Space-based Radio Frequency (RF)}
\label{RF}
RF signals are electromagnetic waves emitted by electronic devices~\cite{unseenlabs-2026-whitepaper}. Ships continuously or intermittently emit such signals through their communication and navigation systems~\cite{pigasse-2024}. While these emissions can be detected by shipborne or shore-based receivers, this study focuses on space-based detections, where satellites in Low Earth Orbit (LEO) capture RF signals from vessels at sea. By analyzing the signals, the geolocation of the source can be determined, even if the transmitting vessel attempts to be invisible~\cite{unseenlabs-2026-whitepaper}.

A key advantage of space-based RF detection is its resilience, as signals can be captured at any time of the day and under all weather conditions~\cite{unseenlabs-2026-whitepaper}. However, a significant limitation is the absence of ship identification. The data is limited to geolocation, timestamp, and signal features~\cite{pigasse-2024}. These features can characterize an emitter across detections, but not reveal which ship it belongs to~\cite{unseenlabs-2026-whitepaper}, necessitating the fusion with other sources (e.g., optical/SAR imagery and AIS)~\cite{morando-2023, srensen-2024}. Another challenge is that vessels typically carry multiple electronic devices, each potentially emitting RF signals simultaneously~\cite{morando-2023}. This can result in multiple detections originating from the same vessel.

\subsection{Pair Hidden Markov Model (PHMM)}
\label{PHMM}
A Pair Hidden Markov Model (PHMM)~\cite{durbin-2005} extends an HMM to supports tasks such as biological sequence alignment and named entity matching~\cite{durbin-2005, nabende-2009}. It takes two input sequences $X = \{x_1, \ldots, x_n\}$ and $Y = \{y_1, \ldots, y_m\}$ over a common alphabet $O$. Moreover, it uses a set of hidden states:
  $B$ (non-emitting start),
  $I_X$ (emits $x_i$, gap in $Y$),
  $I_Y$ (emits $y_j$, gap in $X$),
  $M$ (emits a pair $(x_i, y_j)$) and
  $E$ (non-emitting end).
In a standard PHMM, each emission distribution ($e_{I_X}$, $e_{I_Y}$, $e_M$) sums to 1.

The Forward algorithm sums the weight of all alignments of $X$ and $Y$ over pairs $(i,j)$~\cite{durbin-2005}, with $S$ the set of states:

\begin{enumerate}[leftmargin=*]
    \item Initialization: a matrix $\mathcal{A}$ of size \((n+1) \times (m+1) \times k\) for $k$ states, with $\mathcal{A}_{0,0}(B) = 1$ and all other entries 0 (index 0 is the empty prefix).

    \item Recursion: for each position \((i, j)\) and each emitting state \(c \in \{I_X, I_Y, M\}\),
    \[
        \mathcal{A}_{i,j}(c) = \sum_{p \in S}
        \mathcal{A}_{i', j'}(p) \cdot t_{pc} \cdot e_c(o_{i,j}),
    \]
    with \((i', j') = (i-1, j)\) for \(I_X\), \((i, j-1)\) for \(I_Y\) and \((i-1, j-1)\) for \(M\); \(o_{i,j}\) is a single symbol (for gap states \(I_X\) and \(I_Y\)) or a pair (for state \(M\)).
    \item Termination: the total score is
    \begin{equation}
        P(X, Y \mid \lambda) = \sum_{p \in S}
        \mathcal{A}_{n,m}(p) \cdot t_{p,\textit{end}},
    \label{eq:totalforwardprobphmm}
    \end{equation}
    which is a probability when the model is normalized. In this context, $\lambda$ denotes the model parameters (transition and emission distributions).
\end{enumerate}

\section{Related work}

Because of the inherent limitations of AIS, numerous studies have explored methods to detect falsified AIS messages~\cite{louart-2023} and to augment AIS with complementary data. A common approach involves combining AIS signals with satellite imagery, including optical data~\cite{ballinger-2024} and Synthetic Aperture Radar (SAR)~\cite{rodger-2021, galdelli-2021}. More recently, AIS has also been fused with radio frequency~(RF) detections from both shore-based and space-based systems. Shore-based systems use radar stations that capture reflections of their own transmitted RF signals, while space-based systems rely on satellites that passively collect emissions from ships. Yang et al.~\cite{yang-2022} employed a graph matching-based deep neural network to align AIS data with shore-based radar tracks, whereas Zhang et al.~\cite{zhang-2020} introduced a trajectory association method (D-TRAP) based on spatial proximity and directional consistency to improve matching robustness in multi-ship environments. The effectiveness of space-based RF systems has been demonstrated in case studies from UnseenLabs~\cite{the-maritime-executive-2022} and HawkEye 360~\cite{hawkeye-360-2022}, two commercial providers of satellite-based RF data. 

Although space-based RF has proven its practical value, its fusion with AIS remains a relatively new field of research. Initial studies by Yang et al.~\cite{yang-2024, yang-2025} applied linear interpolation and dead reckoning to temporally align AIS signals with RF detections. Other studies have extended this approach by incorporating SAR imagery as a third data source~\cite{morando-2023, srensen-2024}. In these multi-source data fusion methods, AIS positions are interpolated to align with both the timestamps of RF detections and the acquisition times of SAR imagery. 

One advantage of shore-based radar systems is their highly frequent measurements, typically at intervals of about two seconds~\cite{yang-2022}, which generate dense tracking data that enable accurate matching with AIS data. However, since such systems are generally stationed near ports and coastlines, they are located in areas with heavy traffic and are therefore subject to interference, which increases the likelihood of misclassifications. In contrast, space-based RF systems offer far greater geographic coverage~(up to $300,000$ $\text{km}^2$ per satellite pass~\cite{european-space-agency-2026}), but provide only sparse coverage with long gaps between observations. This significantly complicates temporal and spatial synchronization with AIS.

Nevertheless, space-based RF data offer considerable potential for detecting AIS manipulation on the high seas, where limited surveillance to date increases the likelihood of illegal activities. For this reason, this study focuses on the integration of AIS and space-based RF observations.

While interpolation is straightforward to implement, representing the actual movements of ships, such as changes in course and speed or the intentional deactivation of AIS, is difficult. The longer the interval between AIS signals, the greater the potential error in the estimated positions~\cite{zaman-2023}. For this reason, this study focuses on statistical methods by using a Pair Hidden Markov Model (PHMM) to evaluate entire AIS trajectories, consisting of hundreds or thousands of signals, against individual RF detections using a probabilistic model.

\section{Data Understanding}
\label{sec:understand}
The goal of the \textit{Data Understanding} phase of the CRISP-DM framework is to collect the relevant data and explore its quality and structure. Because commercial providers such as UnseenLabs and HawkEye~360 do not offer public access to RF data, we generate synthetic RF detections from AIS data, based on publicly documented characteristics of real RF systems (Section~\ref{sec:prep}).

The AIS dataset comes from NOAA's Marine Cadastre~\cite{noaa-office-for-coastal-management-2025} and contains transmissions from vessels operating along the United States coastline between 1 and 31 January 2024. It includes roughly \(222\) million AIS messages sent by \(33{,}635\) unique vessels, grouped into \(83\) vessel categories such as fishing, military, and cargo ships. Each AIS message contains identification fields (e.g., MMSI and IMO number), position, speed, and course. The presence of unique identifiers allows to group messages by vessel and reconstruct trajectories, defined as continuous sequences of AIS messages for a given ship between a start and end time. 

Pleasure crafts, yachts, and fishing vessels form the largest share of trajectories, but they mainly use Class B AIS transponders which transmit less frequently than Class A and lead to sparser and less accurate tracks (\(91\%\) of pleasure crafts, \(75\%\) of fishing vessels). In contrast, \(99\%\) of cargo vessels use Class A transponders. For this reason, the analysis focuses on cargo vessels, yielding \(2{,}400\) trajectories.

An exploratory analysis of these cargo vessel trajectories produced three main insights. First, more than \(600\) cargo vessels transmit between \(5{,}000\) and \(10{,}000\) messages per voyage, while a small group records fewer than five messages. This likely indicates incomplete records or very short trips. Second, the most frequently observed mean time interval between consecutive AIS messages falls between \(150\) and \(180\)~seconds ($2.5$-$3$~minutes). This is longer than the nominal Class A reporting intervals ($2$ to $10$~seconds underway, $3$~minutes at anchor or moored), possibly due to resampling or limited reception. Many trajectories, however, contain mean intervals above \(600\) seconds ($10$~minutes), which can point to data quality issues, equipment problems, or deliberate AIS deactivation. Third, the vessel speeds, computed from the geographic distance and time between successive AIS messages, are for most cargo vessels between \(5\) and \(20\)~km/h. This is plausible given the port and coastal segments in the data. A considerable share of the trajectories averages below \(5\)~km/h and some contain very high speeds, so both groups merit closer inspection in later preprocessing.

\section{Data preparation}
\label{sec:prep}

The \textit{Data Preparation} stage involves cleaning and structuring of the AIS data as a time-ordered sequence suitable for use in a PHMM and generating synthetic RF data from the processed AIS dataset.

\subsection{AIS Data Preprocessing Pipeline}
\label{preprocessing}

As the RF detections are generated synthetically and anchored to AIS tracks, the AIS data serves as ground truth for the matching task. It must therefore contain only plausible, continuously transmitted vessel behavior. Implausible kinematics or transmission gaps would make the ground-truth label of a synthetic RF point ambiguous, regardless of whether their underlying cause is a data-quality artifact or a dark vessel. The raw cargo AIS data ($2,400$ unique cargo trajectories) is accordingly preprocessed with a multi-step pipeline that removes such cases and structures the remaining data for sequence alignment. Four steps were applied.

\paragraph{Step 1: Removal of Short Trajectories}

The first step is to remove trajectories with five or fewer AIS messages. The aim is to retain only trajectories with sufficient information value and a reasonable likelihood of temporal alignment with synthetic RF detections, which are typically acquired at relatively long intervals. In total, $80$ trajectories were excluded, leaving $2,320$ for subsequent filtering. As a result, the average number of AIS messages per trajectory increased by $3.45\%$, rising from approximately $6,117$ to $6,328$.

\paragraph{Step 2: Removal of Speed Outliers}

The second step addresses outliers in vessel speed values. First, trajectories with unrealistically high-speed samples are removed using the interquartile range (IQR) method, with trajectories exceeding $Q3 + 1.5 \times (Q3 - Q1)$ considered outliers. This aims to preserve trajectories exhibiting plausible movement patterns. Second, trajectories with an average speed below $5$ km/h are removed, as they are assumed to represent stationary vessels not considered in this study.

In total, $952$ trajectories (high-speed outliers and trajectories with an average speed below $5$~km/h) were excluded, leaving $1,368$ trajectories. The majority of cargo vessels maintained average speeds between $7$ and $20$~km/h, with a mean of $14.03$~km/h. Only a few trajectories exhibited average speeds above $35$~km/h, with a maximum of $40$~km/h, which still is a plausible operating speed under certain conditions \cite{wang-2012}. 

\paragraph{Step 3: Segmentation of Trajectories}

The third step addresses substantial temporal gaps in trajectories. Trajectories containing gaps longer than 551 seconds (approximately nine minutes, corresponding to the $99$th percentile of all inter-message time gaps pooled across the Step 2 output) are segmented into shorter trajectories. The aim is to enhance continuity and minimize interruptions caused by extended periods without recorded positions. Each newly created segment is additionally subject to the same minimum-length (Step 1) and minimum-average-speed (Step 2) criteria. The segmentation increased the dataset size from $1,368$ to $16,727$ trajectories. Segments of the same vessel are treated as distinct candidates. Since the temporal preselection (Sec.~\ref{preselection}) requires a candidate to cover the RF timestamp, and segments split at gaps are disjoint in time, only $10$ of $17,081$~RF points have another segment of the true vessel in their candidate set, so this has a negligible effect on the reported precision.

\paragraph{Step 4: Sampling and Sequence Preparation}

The fourth step prepares the trajectories for subsequent analysis. This step aims to reduce computational time by creating a smaller yet representative dataset while preserving trajectory diversity. To this end, a sample of $5,000$ trajectories is selected using Kernel Density Estimation (KDE). Rather than uniform random sampling, KDE assigns selection probabilities based on selected features of an AIS track (total number of AIS messages, voyage duration, average speed, geographical displacement latitude and longitude, average time interval between consecutive messages). This ensures a balance between common and rare trajectory patterns. Each trajectory in this sample is then converted into a sequence of tuples $(\texttt{timestamp}_{\texttt{UTC}}, (\texttt{latitude}, \texttt{longitude}))$ and ordered chronologically to maintain temporal consistency.

\subsection{RF Data Generation}
\label{rf-generation}

In the absence of publicly available RF data, synthetic RF detections were generated from the AIS data. To enable PHMM-based sequence alignment, these detections adopted the same data structure as the AIS records, with each RF detection represented as a tuple $(\texttt{timestamp}_{\texttt{UTC}},(\texttt{latitude}, \texttt{longitude}))$. Unlike AIS sequences, which typically contain hundreds to thousands of observations, each RF sequence comprises only a single detection, consistent with the study's objective of associating individual RF observations with complete AIS trajectories. 

Commercial specifications (e.g., UnseenLabs) indicate two defining characteristics of current satellite constellations: (1) a revisit time of approximately three hours (the interval between consecutive detections), and (2) a geolocation accuracy of roughly one kilometer (km) under optimal conditions~\cite{european-space-agency-2026}. Based on these characteristics, the process was organized into three sequential steps, applied to the $5,000$ AIS trajectories after preprocessing. On average, this procedure generated 3.42 synthetic RF detections per AIS trajectory in the dataset.

\paragraph{Step 1: Build Sampling Schedule}
First, a 24-hour sampling schedule is constructed to mimic the temporal characteristics of satellite observations. A random initial timestamp is drawn from a predefined time window, after which subsequent timestamps are generated at three- to four-hour intervals with uniform distribution. Note that all generated RF-points will sample from the same schedule, replicating the revisit patterns of UnseenLabs satellites.

\paragraph{Step 2: Align RF Timestamps to AIS Trajectories}
The second step aligns the simulated RF timestamps with the temporal range of AIS trajectories. To allow for detections slightly outside the original trajectory boundaries, this range is extended by a fixed buffer (five minutes at both the start and the end). RF timestamps that fall within the extended window are then matched to the closest AIS message in time.

\paragraph{Step 3: Generate Synthetic RF Coordinates}
Finally, synthetic RF coordinates are generated by introducing noise to the associated AIS position. AIS positions are offset by sampling from a two-dimensional (east, north) normal distribution $\mathcal{N}(0,\sigma)$ with $\sigma=800 \text{m}$ to simulate satellite geolocation uncertainty. The RF position is thus the position of the closest AIS message plus this offset, and vessel movement between the AIS and RF timestamps is not modeled. Assuming that north- and east errors are uncorrelated and have the same variance, then the magnitude of the error follows a Rayleigh distribution. Then, the mean error is $\mu=\sigma\sqrt{\pi/2}\approx 1000m $, reflecting UnseenLabs’ reported geolocation accuracy of approximately one kilometer under optimal conditions.

\section{Modeling}
\subsection{PHMM}
\label{sec:phmm}
\begin{figure}
    \centering
    \includegraphics[width=\columnwidth]{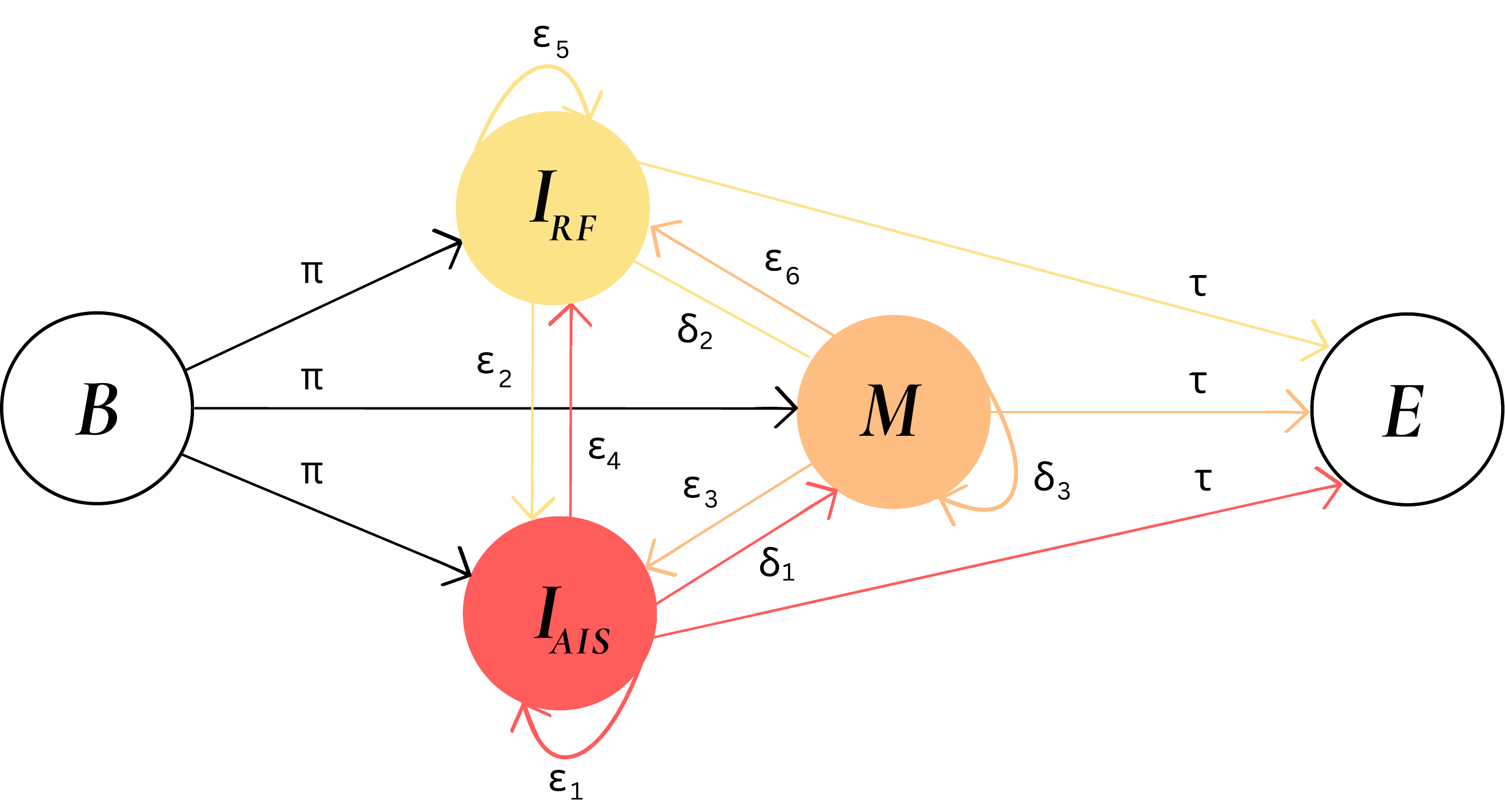} 
    \caption{Graphical representation of the PHMM.}
    \label{fig:architecture}
\end{figure}

This section addresses the \textit{Modeling} phase of the CRISP-DM framework by formally defining the PHMM. The two input sequences $X$ and $Y$ of the PHMM are $AIS$ (AIS messages from a candidate trajectory), and $RF$ (the RF detection to be associated with that trajectory). $AIS$ has $n$ messages and $RF$ consists of a single detection, so the length of the second sequence is $m=1$. The respective state architecture (cf. transition architecture in Figure~\ref{fig:architecture}) therefore consists of $B$ (begin), $E$ (end), $M$ (match), $I_{AIS}$ (insert in $AIS$), and $I_{RF}$ (insert in $RF$) states. We define each observation $o$ as a tuple $(\texttt{timestamp}_{\texttt{UTC}}, (\texttt{latitude}, \texttt{longitude}))$, where values are unconstrained within their natural bounds.

\begin{table}[t]
\centering
\caption{PHMM transition parameters. $\pi$ and $\tau$ are 
uniform values; $\varepsilon$ and $\delta$ are hand-designed
functions of $\Delta_t$ and $\Delta_d$.}
\label{tab:transitionmatrix}

\setlength{\tabcolsep}{3pt}
\renewcommand{\arraystretch}{1.15}

\scalebox{0.8}{%
\begin{tabular}{@{}
    p{0.18\columnwidth}
    p{0.60\columnwidth}
    p{0.40\columnwidth}
@{}}
\toprule
Matrix element & Explanation & Value/distribution \\
\midrule

$\pi$
& Initial state distribution: transition from $B$ to
an emitting state ($I_{AIS}$, $I_{RF}$, $M$).
& Uniform, $1/3$. \\

$\tau$
& Transitions from an emitting state
($I_{AIS}$, $I_{RF}$, $M$) to termination state $E$.
& Uniform, $0.05$. \\

$\varepsilon$
& Transitions into $I_{AIS}$ and $I_{RF}$.
& $f_{\varepsilon}(\Delta_t,\Delta_d)$,
see Figure~\ref{fig:AIS-RF-transitions}. \\

$\delta$
& Transitions into $M$.
& $f_{\delta}(\Delta_t,\Delta_d)$,
see Figure~\ref{fig:M transitions}. \\

\bottomrule
\end{tabular}%
}
\end{table}

Transition weights specify the strength of moving from one hidden state to another during sequence alignment. While classical (P)HMMs assume stationary probabilities independent of the observations, we model the transition matrix $\mathcal{T}$ as depending on the spatio-temporal context (cf. Table~\ref{tab:transitionmatrix}):

\vspace{-2em}
\begin{equation}
  \mathcal{T} =
  \begin{array}{@{}c@{}}
    \rowind{B} \\ \rowind{$I_{AIS}$} \\ \rowind{$I_{RF}$} \\ \rowind{M} \\ \rowind{E}
  \end{array}
  \mathop{\left[
  \begin{array}{ *{5}{c} }
     \colind{0}{B}  & 
     \colind{\pi}{$I_{AIS}$}  &  
     \colind{\pi}{$I_{RF}$} & 
     \colind{\pi}{M} & 
     \colind{0}{E}\\
    0  & {\epsilon_1}  & {\epsilon_4} & {\delta_1} & {\tau}\\
    0 & {\epsilon_2}  & {\epsilon_5}  & {\delta_2} & {\tau}\\
    0  & {\epsilon_3} & {\epsilon_6} & {\delta_3} & {\tau}\\
     0  & 0  & 0 & 0 & 0\\
  \end{array}
  \right]}^{
  \begin{array}{@{}c@{}}\\ \mathstrut
  \end{array}
  }
  \label{eq:transmatrixstudy}
\end{equation}

Each $\epsilon_\cdot$ is $f_\varepsilon$ and each $\delta_\cdot$ is $f_\delta$, evaluated at the corresponding pair of observations. With $m=1$, the transitions $\epsilon_5$, $\epsilon_6$, $\delta_2$ and $\delta_3$ cannot occur, since they would require a second RF symbol. The remaining weights depend on the time difference $\Delta_t$ (in seconds) and the haversine distance $\Delta_d$ (in kilometers) between consecutive observations. For the $I_{AIS}$ and $I_{RF}$ transitions ($f_\varepsilon$), negative $\Delta_t$ values (i.e., backward time transitions) are assigned zero weight to enforce chronological ordering. The $M$-state function $f_\delta$ instead depends on the absolute difference $|\Delta_t|$ and therefore does not distinguish a forward from a backward AIS--RF pairing. Both functions are hand-designed. The cut-offs of $120$\,s and $2$\,km in $f_\delta$ were set with knowledge of the generator ($\sigma=800$\,m, Sec.~\ref{rf-generation}).

    \begin{figure}[htp]
        \centering
        \includegraphics[width=0.9\linewidth]{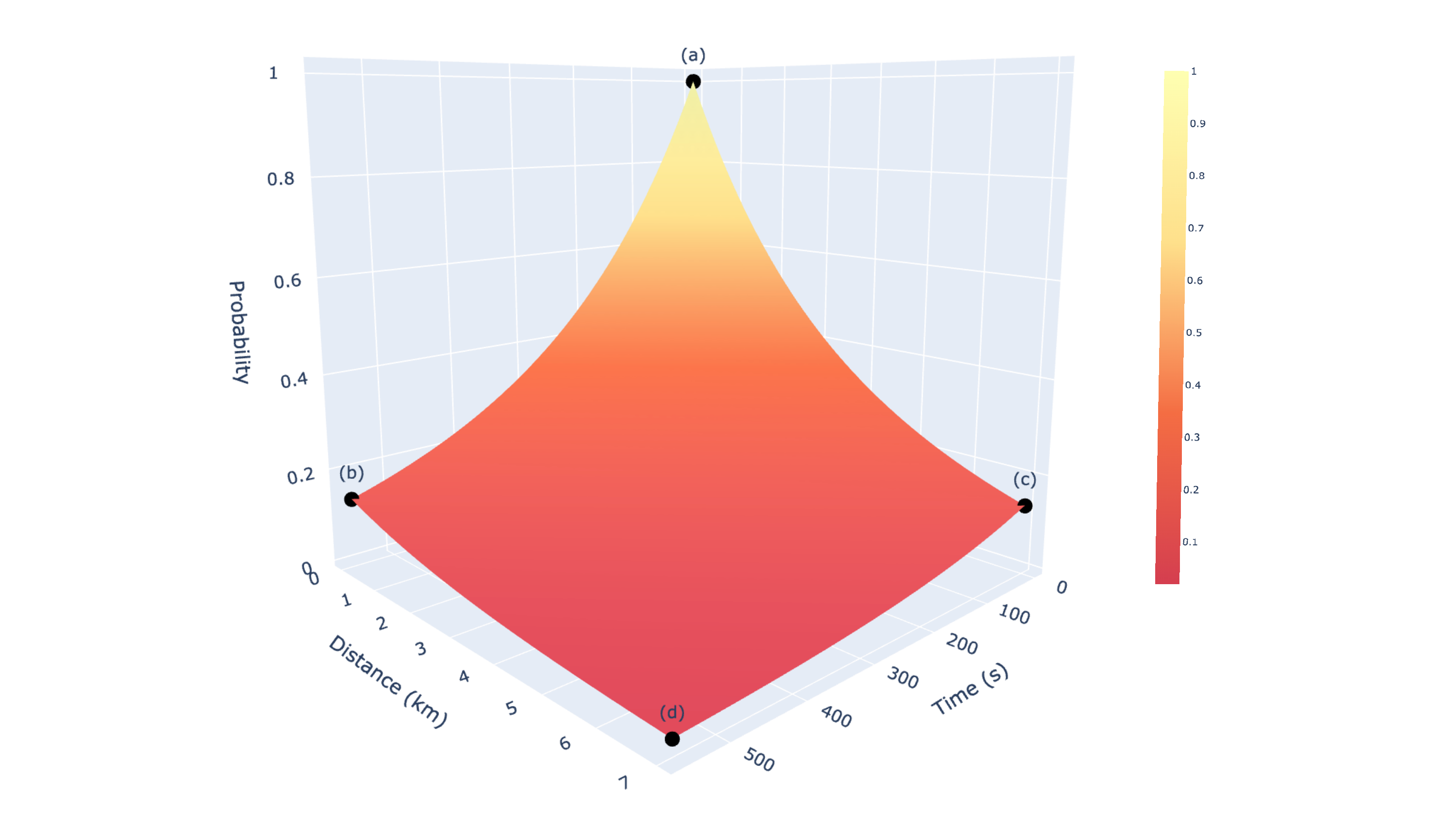} 
        \caption{Weight surface of $f_{\varepsilon}(\Delta_t, \Delta_d)$. This function attains its maximum value for observations that occur nearly simultaneously and at the same location (point~\textit{a}), suggesting a high probability that they originate from the same vessel. The probability decays sharply with increasing temporal (point~\textit{b}) or spatial (point~\textit{c}) separation, reaching its minimum when both differences are substantial (point~\textit{d})}
        \label{fig:AIS-RF-transitions}
    \end{figure}

    \begin{figure}[htp]
        \centering
        \includegraphics[width=0.9\linewidth]{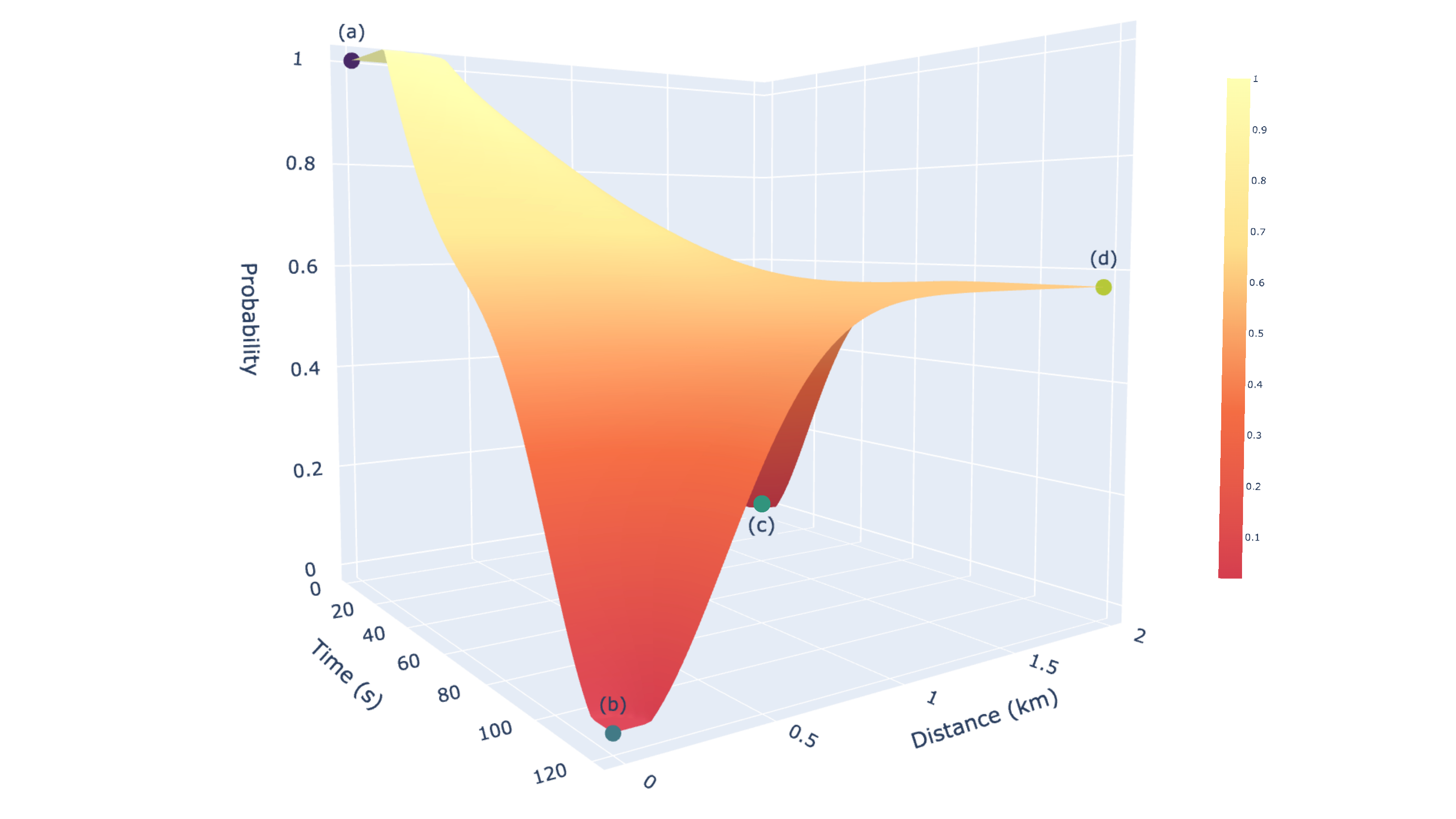} 
        \caption{Weight surface of $f_\delta(\Delta_t, \Delta_d)$. This function is zero beyond predefined temporal ($120$ seconds) and spatial ($2$ kilometers) thresholds, so pairs outside these thresholds cannot transition into the $M$ state}
        \label{fig:M transitions}
    \end{figure}

At each alignment step, the transition weights from the previous state to all possible current states are scaled to sum to $0.95$, with the remaining $0.05$ allocated to transitions to the termination state ($\tau$). An exponent $\kappa=(\kappa_{AIS},\kappa_{RF})=(0.5,2)$ is then applied to these weights, with $\kappa_{AIS}$ for transitions into an AIS state and $\kappa_{RF}$ for transitions into $I_{RF}$ or $M$. These values were set by informal exploration, possibly on data overlapping the evaluation set, and $\alpha$ (Sec.~\ref{lengthcorrection}) was tuned conditional on them. The rows of $\mathcal{T}$ therefore no longer sum to $1$, and we call the model PHMM-style.

With $\kappa=(1,1)$ and $\alpha$ re-tuned, the length-corrected Forward score drops from $0.7776$ to $0.6934$ in precision and from $0.637$ to $0.556$ in AUC, while the two log-odds variants change by at most $0.0009$ and $0.0063$. Part of the Forward baseline's performance is thus due to $\kappa$.

The emission weights define what an $I_{AIS}$ or $I_{RF}$ state emits for a single symbol or what an $M$ state emits for a pair of aligned symbols. $I_{AIS}$ and $I_{RF}$ states emit the respective $\mathrm{ais}_i$ and $\mathrm{rf}$ symbols with weight 1. In the $M$ state, a pair $(\mathrm{ais}_i, \mathrm{rf})$ is emitted with a weight given by the degree of spatio-temporal correspondence between $\mathrm{ais}_i$ and $\mathrm{rf}$, defined by the function $f_{\delta}(\Delta_t, \Delta_d)$. The function $f_\delta$ therefore enters twice, as the weight of the transition into $M$ and as the (normalized) emission weight of $M$, with $\kappa_{RF}=2$ applied to the transition weight.

\subsection{AIS-RF Preselection}
\label{preselection}
Many AIS-RF combinations are unlikely to align meaningfully due to clear temporal and spatial discrepancies, so a preselection procedure is applied. The number of candidate pairs after preselection is identical for all methods. Each RF detection is compared against AIS trajectories using three successive filtering steps. Pairs that satisfy all three criteria are retained as a candidate dataset. 

First, we employ a \textit{temporal filtering} step: the timestamp of an RF detection must fall within the temporal window of a given AIS trajectory, defined as the interval between the first and last AIS timestamps, extended by ten minutes at both ends to account for small temporal deviations.

Second, \textit{bounding box filtering} limits the location of the RF detection to be included within the trajectory's bounding box, determined by the minimum and maximum latitude/longitude of all AIS points in the trajectory, expanded by 6 km in all directions to account for small spatial deviations.

Third, the \textit{proximity filtering} restricts the relevant AIS trajectories to messages recorded within ± $3$ hours of the RF detection timestamp. The RF-AIS pair is retained only if at least one AIS point lies within a $6$ km radius of the RF detection's location. Both the ± $3$-hour temporal window and the $6$ km spatial threshold follow the methodology outlined in prior work~\cite{yang-2024}.

\subsection{Alignment via Forward Algorithm}
\label{proposedforward}

For each RF detection, the Forward algorithm of Sec.~\ref{PHMM} is applied to every candidate AIS trajectory, with $X$ the trajectory ($n$ messages, varying per candidate), $Y$ the single RF detection ($m=1$) and $k=5$ states. For numerical stability the recursion runs in log space. All cells are initialized to $-\infty$ ($\log 0$) except $\mathcal{A}_{0,0}(B)=0$, and the sum over predecessor states becomes a log-sum-exp (LSE):

\vspace{-10pt}
{\small
\begin{equation}
\begin{split}
\mathcal{A}_{i,j}(c) = \operatorname{LSE}_{p\in\mathrm{Pred}(c)}\Big(
  &\mathcal{A}_{i',j'}(p) \\
  &+ \log t_{pc} + \log e_c(o_{i,j})\Big).
\end{split}
\label{eq:pair_forward_log}
\end{equation}
}

Termination (Eq.~\ref{eq:totalforwardprobphmm}) yields the raw Forward score $\log P_{\mathrm{full}}=\operatorname{LSE}_p\big(\mathcal{A}_{n,1}(p)+\log t_{p,E}\big)$, and the candidate with the highest score is predicted. This value is not a probability, because the $I$-state emissions are 1, $f_\delta$ is a bounded similarity function, the transitions depend on the observations and are row-scaled to 0.95, and $\kappa$ is applied after normalization, so the rows of $\mathcal{T}$ no longer sum to 1.

\paragraph{Log-odds score.} With $m=1$, the AIS$\to$AIS transition weights before row scaling do not depend on the RF point. The raw score therefore mostly reflects the AIS trajectory itself (a trajectory prior) and contains no null hypothesis. The RF point enters through the paths that visit $I_{RF}$ or $M$ and, via the row scaling, through the AIS$\to$AIS transitions.

We therefore also score each candidate by the log-odds of the full model against a null model, as in profile HMMs~\cite{durbin-2005} (a log-likelihood ratio),
\begin{equation}
s = \log P_{\mathrm{full}} - \log P_{\mathrm{null}},
\label{eq:logodds}
\end{equation}
where $P_{\mathrm{null}}$ is the Forward score of a null model. The null model has the same transitions as the full model, except that every transition into $M$ is set to zero after the row scaling to 0.95 and the exponent $\kappa$ have been applied. The other transition weights are left unchanged. No path of the null model can visit $M$, so $s \ge 0$.

This null model differs from the random model of Durbin et al.~\cite{durbin-2005}, which has its own emission distribution. The fraction of the Forward weight that lies on paths through $M$, $1-e^{-s}$, increases monotonically with $s$, so ranking by it gives the same ranking and ROC curve as $s$. We use raw $s$ as the primary variant and $s/n^{\alpha}$ (Sec.~\ref{lengthcorrection}) as a secondary variant.

\subsection{Length correction}
\label{lengthcorrection}
The log Forward score decreases with trajectory length. Scores are therefore divided by $n^{\alpha}$, where $n$ is the number of AIS messages:

{\small
\begin{equation}
\mathrm{score}_{\alpha} =\frac{\log P_{\mathrm{full}}}{n^{\alpha}}.
\label{eq:normalized_log_prob}
\end{equation}
}

$\alpha=1$ is the length-normalized score; a tuned $\alpha$ gives the length-corrected score. $\alpha$ is chosen for precision on a $20\%$ tuning split and evaluated on the remaining $80\%$ validation split. Applied to the log-odds score, the same correction ($s/n^{\alpha}$) is a secondary variant, with $\alpha$ tuned separately for $s$ on the same tuning split ($\alpha=1.27$).

\subsection{Nearest-Neighbor Baselines}
\label{sec:baselines}

Since the suitability of a method depends on trade-offs such as accuracy and computational cost~\cite{Pistorius.2020}, we compare the PHMM scores against two nearest-neighbor (NN) baselines on the same pre-filtered candidate set. For an RF observation at position $p_{\mathrm{RF}}$ and time $t_{\mathrm{RF}}$, and a candidate AIS track $T = (s_1, \dots, s_n)$ with timestamps $t_1, \dots, t_n$, each baseline reduces $T$ to one distance, and the predicted match is the candidate that minimizes it.

The geometric baselines take the minimum distance from $p_{\mathrm{RF}}$ to the track points (Euclidean on latitude/longitude degrees, or Haversine great-circle) or to the track's $n{-}1$ consecutive-point segments (point-to-segment, in a local flat-earth plane centered on $p_{\mathrm{RF}}$). For segment $i$, $d_i$ is the distance from $p_{\mathrm{RF}}$ to its closest point $s_i + t^*(s_{i+1}-s_i)$, where
\begin{equation}
t^* = \mathrm{clip}\!\left(\frac{(p_{\mathrm{RF}}-s_i)\cdot(s_{i+1}-s_i)}{\|s_{i+1}-s_i\|^2},\,0,\,1\right)
\end{equation}
is its fractional position along the segment. All three were tried and differ little, by at most $0.0081$ in precision and $0.0033$ in AUC. We report point-to-segment, which has the highest AUC of the three but the lowest precision.

The time-weighted baseline adds a time term at the interpolated AIS timestamp $\hat{t}_i = t_i + t^*(t_{i+1}-t_i)$:
\begin{equation}
  d_{\mathrm{tw}}(p_{\mathrm{RF}}, T) = \min_{i} \left[
    \frac{d_i}{d_{\mathrm{thr}}} + w \, \frac{|t_{\mathrm{RF}} - \hat{t}_i|}{T_w}
  \right],
\end{equation}
where $d_{\mathrm{thr}} = 6$\,km and $T_w = 3$\,h normalize the two terms to the scale of the preselection window whereas $w = 1$ weights them equally.

\section{Evaluation}
\subsection{Experimental Setup}
Table~\ref{tab:populations} lists the data used in this section. The $5{,}000$ sampled trajectories yield $17{,}081$ RF detections. For $10{,}903$ of them (single-candidate) the prefilter leaves only the true trajectory as candidate, so the match is trivial. The other $6{,}178$ are multi-candidate detections with two to nine candidates (usually two or three). Only these detections discriminate between scores.

The length-correction exponent $\alpha$ is tuned on a random $20\%$ of these ($1{,}236$ detections) and precision is reported on the remaining $80\%$ ($4{,}942$). The nearest-neighbor comparison and the bias experiments use the same $4{,}942$ detections. Precision is top-1 accuracy. Each detection yields one prediction, the highest-scoring candidate, so precision equals Recall@$1$. Recall@$3$ and Recall@$5$ are above $0.98$ and $0.99$ for every score, so we report only precision.

The dark-vessel test evaluates whether a score recognizes detections from vessels without an AIS trajectory. The $3{,}203$ trajectories that are the source or a candidate of at least one detection (out of the $5{,}000$ sampled) belong to $1{,}022$ vessels, identified by their MMSI. We split these vessels at random into a known group ($80\%$) and a dark group ($20\%$), which keeps all trajectories of a vessel together. Only the $2{,}517$ trajectories of known vessels are offered as candidates. The $13{,}751$ detections from known vessels have their true trajectory among the candidates and form the positive class. The $3{,}330$ detections from dark vessels have no correct candidate and form the negative class. A score separates the classes if the score of the top-ranked candidate is high for positives and low for negatives.

We report the AUC of this separation, which needs no threshold. The spatio-temporal prefilter already rejects $2{,}346$ ($70.45$\%) of the dark detections because no candidate is close enough. They are left out of the AUC, which is computed on the positives and the remaining $984$ dark detections, where the decision depends on the score alone. The AUC reuses the $\alpha$ tuned for precision (Sec.~\ref{lengthcorrection}) without re-tuning and applies it to all detections.

\begin{table}[t]
\caption{Evaluation populations. Top: RF detections used for top-1 precision, the bias experiments and the baseline comparison, with the precision of PHMM Forward at each stage. Bottom: vessels (MMSIs), trajectories and RF detections used for the dark-vessel test.}
\label{tab:populations}
\centering
\begin{tabular}{@{}lrl@{}}
\toprule
RF detections & $n$ & Precision \\
\midrule
All & 17{,}081 & 0.8899\textsuperscript{a} \\
\quad Single-candidate & 10{,}903 & 1.0000 \\
\quad Multi-candidate & 6{,}178 & 0.6955\textsuperscript{a,b} \\
\quad\quad Tuning split (20\%) & 1{,}236 & 0.7921 ($\alpha=0.90$) \\
\quad\quad Validation split (80\%) & 4{,}942 & 0.6938 $\rightarrow$ 0.7776\textsuperscript{c} \\
\bottomrule
\end{tabular}

\medskip
\begin{tabular}{@{}lrrr@{}}
\toprule
Dark-vessel test & MMSIs & Trajectories & Detections \\
\midrule
Known (positives) & 818 & 2{,}517 & 13{,}751 \\
Dark (negatives) & 204 & 686 & 3{,}330 \\
All & 1{,}022 & 3{,}203 & 17{,}081 \\
\bottomrule
\end{tabular}

\vspace{2pt}
\footnotesize\textsuperscript{a}length-normalized ($\alpha=1$, Sec.~\ref{lengthcorrection}). The tuned $\alpha=0.90$ correction is fit and evaluated only on the tuning and validation splits below, not on this row. \textsuperscript{b}Multi-candidate = tuning split + validation split (1{,}236 + 4{,}942 = 6{,}178). \textsuperscript{c}Before $\rightarrow$ after length correction.
\end{table}

\subsection{Results}
\paragraph{Precision}
PHMM Forward's precision falls from $0.8899$ on the full dataset to $0.6955$ once restricted to multi-candidate detections, and length correction raises it from 0.6938 to $0.7776$ on the validation split (Table~\ref{tab:populations}). The log-odds score reaches $0.8580$ on the same split ($0.8571$ with $n^{\alpha}$). For raw log-odds, $203$ of the $4{,}942$ validation detections have several top-scoring candidates. Ties are resolved by selecting the candidate with the lowest trajectory ID.

\paragraph{Bias Experiments}
For the detections a score gets wrong, we compare the trajectory it ranks highest (\textit{chosen}) with the true trajectory (\textit{should-be}). The \textit{correct} group contains the detections where the true trajectory ranks highest. We test whether the score favors longer trajectories (Experiment 1), densely sampled trajectories (Experiment 2, gaps between consecutive AIS positions), stationary vessels (Experiment 3, share of AIS messages below $5$\,km/h) or candidates close to the RF detection in time and space (Experiment 4). Table~\ref{tab:bias} reports the results for the corrected Forward score and the log-odds score.

\begin{figure*}[t]
\centering
\begin{subfigure}[t]{0.32\textwidth}
    \centering
    \includegraphics[width=\linewidth]{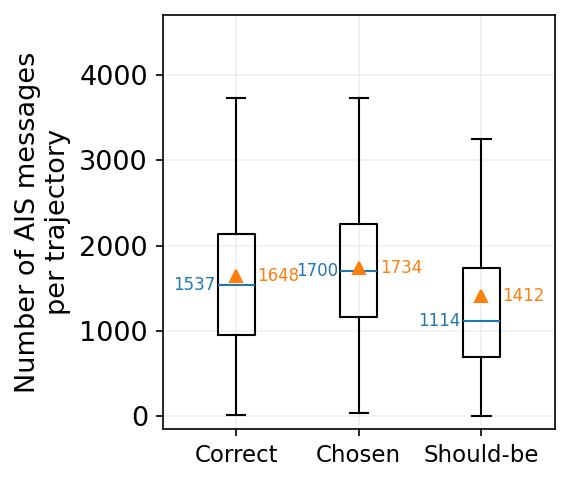}
    \caption{PHMM Forward, corrected}
    \label{fig:tracklen-forward}
\end{subfigure}\hfill
\begin{subfigure}[t]{0.32\textwidth}
    \centering
    \includegraphics[width=\linewidth]{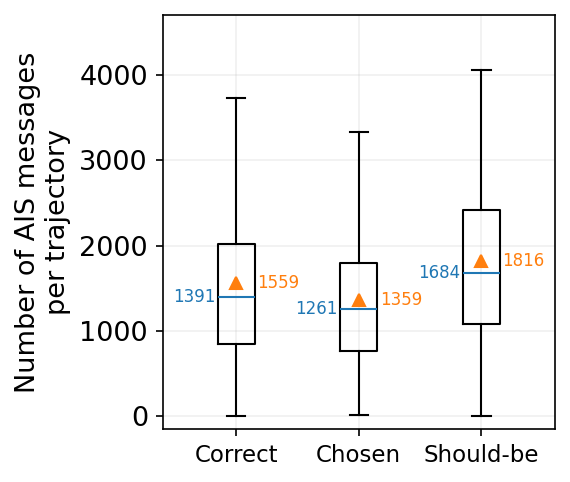}
    \caption{PHMM log-odds ($n^{\alpha}$ variant)}
    \label{fig:tracklen-logodds}
\end{subfigure}\hfill
\begin{subfigure}[t]{0.32\textwidth}
    \centering
    \includegraphics[width=\linewidth]{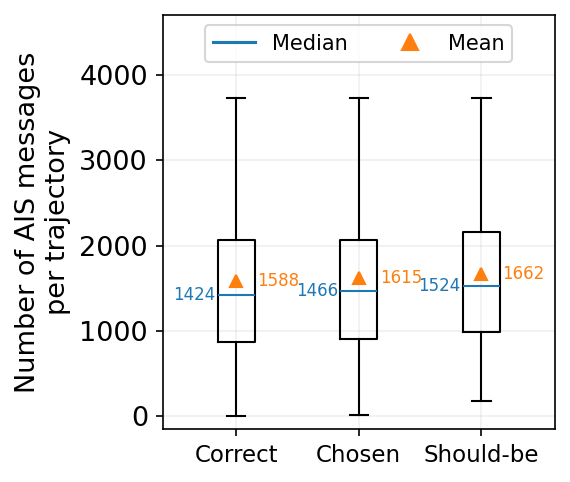}
    \caption{NN time-weighted}
    \label{fig:tracklen-nn}
\end{subfigure}
\caption{AIS messages per trajectory for the Correct, Chosen and Should-be groups. The numbers next to the boxes are the means (orange) and medians (blue), and the legend in (c) applies to all panels.}
\label{fig:tracklen}
\end{figure*}

For corrected Forward, chosen trajectories are longer than should-be ($p=2.2\times10^{-14}$). The correction reduces this bias but does not remove it ($t$ = $27.96$ before, $7.70$ after). They have shorter gaps between consecutive AIS positions but not shorter time gaps, and are more stationary. Their closest AIS point is also farther from the RF detection, in both time and distance. The bias tests for the log-odds score were run on the $n^{\alpha}$ variant only (Fig.~\ref{fig:tracklen}). There, the differences in intra-track distance gap, stationarity and AIS--RF time gap are no longer significant, while the AIS--RF distance gap difference persists ($p=3.8\times10^{-14}$, versus $5.9\times10^{-30}$). Chosen trajectories are now shorter than should-be ($p=1.0\times10^{-20}$) and have larger intra-track time gaps ($p=0.040$, uncorrected for multiple tests).

\begin{table}[t]
\caption{Bias experiments, validation split. Cells give the direction (chosen vs.\ should-be) and the two-sided $p$-value where the test is significant at $p<0.05$ (Welch for 1--3, paired for 4a--4b), and n.s.\ otherwise. The log-odds column is the $n^{\alpha}$ variant. NN time-weighted (not shown) has no significant bias except in the AIS--RF time gap (4a), which is expected, since its score contains the time difference directly.}
\label{tab:bias}
\centering
\footnotesize
\setlength{\tabcolsep}{3pt}
\begin{tabular}{@{}lll@{}}
\toprule
Experiment & \makecell[l]{Forward\\(corrected)} & \makecell[l]{Log-odds\\($n^{\alpha}$)} \\
\midrule
1. Track length & longer, $2.2{\times}10^{-14}$ & shorter, $1.0{\times}10^{-20}$ \\
\makecell[l]{2a. Intra-track\\time gap} & n.s. & larger, $0.040$ \\
\makecell[l]{2b. Intra-track\\distance gap} & closer, $7.2{\times}10^{-8}$ & n.s. \\
3. Stationarity & higher, $2.0{\times}10^{-5}$ & n.s. \\
4a. AIS--RF time gap & farther, $1.8{\times}10^{-3}$ & n.s. \\
4b. AIS--RF distance gap & farther, $5.9{\times}10^{-30}$ & farther, $3.8{\times}10^{-14}$ \\
\bottomrule
\end{tabular}
\end{table}

\begin{figure}[t]
\centering
\includegraphics[width=\linewidth]{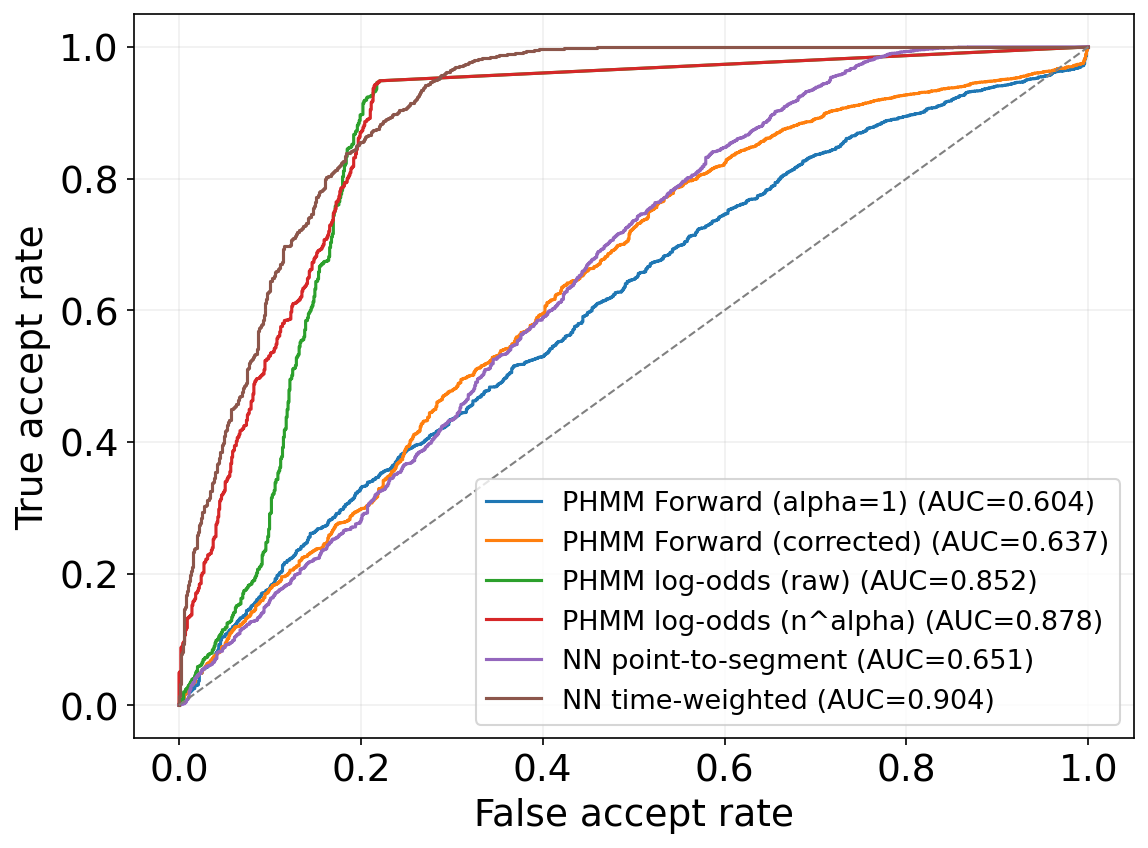}
\caption{ROC of the dark-vessel test on the detections not rejected by the prefilter, for PHMM Forward ($\alpha=1$ and corrected), PHMM log-odds (raw and $n^{\alpha}$), NN point-to-segment and NN time-weighted. Legend AUC values match Table~\ref{tab:main}.}
\label{fig:openset}
\end{figure}

\paragraph{Baseline Comparison}
In Table~\ref{tab:main}, precision is measured on the validation split and the AUC on the dark-vessel test (Fig.~\ref{fig:openset}). The log-odds score raises the AUC from 0.637 (corrected Forward) to 0.852, and to 0.878 with $n^{\alpha}$. Corrected Forward exceeds NN point-to-segment in precision (0.7776 vs.\ 0.6333) but not in AUC (0.637 vs.\ 0.651). Both log-odds variants exceed it on both metrics. NN time-weighted requires no fitting and combines distance with a linear time penalty. It outperforms the PHMM scores on both metrics (precision 0.9013 vs.\ 0.8580, AUC 0.904 vs.\ 0.852 and 0.878 for raw and $n^{\alpha}$). For precision, the difference between raw log-odds and NN time-weighted is significant (McNemar $\chi^2=65.9$, $p<10^{-15}$). NN time-weighted alone is correct on 451 detections and raw log-odds alone on 237. For the $n^{\alpha}$ variant, $\chi^2=69.0$ and $p<10^{-15}$. The AUC gap is reported without a significance test.

\begin{table}[t]
\caption{Precision (validation split, $n=4{,}942$) and AUC of the dark-vessel test (detections not rejected by the prefilter).}
\label{tab:main}
\centering
\begin{tabular}{@{}lrr@{}}
\toprule
Model & Precision & AUC (dark-vessel) \\
\midrule
PHMM Forward ($\alpha=1$) & 0.6938 & 0.604 \\
PHMM Forward (corrected) & 0.7776 & 0.637 \\
PHMM log-odds (raw)      & 0.8580 & 0.852 \\
PHMM log-odds ($n^{\alpha}$) & 0.8571 & 0.878 \\
NN point-to-segment\textsuperscript{a} & 0.6333 & 0.651 \\
NN (time-weighted) & \textbf{0.9013} & \textbf{0.904} \\
Random baseline    & 0.4141 & -- \\
\bottomrule
\end{tabular}

\vspace{2pt}
\footnotesize\textsuperscript{a}Selected from Euclidean, Haversine and point-to-segment (Sec.~\ref{sec:baselines}). The three differ by at most 0.0081 in precision and 0.0033 in AUC. The random baseline is the expected top-1 precision when a candidate is chosen uniformly at random.
\end{table}

\section{Discussion}
\subsection{Experimental Outcomes}
With a single RF detection, the raw Forward score is largely a trajectory prior (Sec.~\ref{proposedforward}). The errors of the corrected Forward score reflect trajectory-level characteristics. It favors longer, more stationary trajectories with shorter inter-message distances, and the closest AIS point of its chosen candidate is farther from the RF detection in both time and distance. Length correction ($\alpha=0.90$) reduces the length bias but does not remove it. In the $n^{\alpha}$ variant, the log-odds score no longer shows significant differences in intra-track distance gap, stationarity and AIS--RF time gap. The raw log-odds score raises precision from 0.7776 to 0.8580 and AUC from 0.637 to 0.852, but NN time-weighted stays ahead on both metrics (0.9013 and 0.904). Because the model structure was not varied, the effect of the scoring rule cannot be compared with the effect of the structure. The evidence supports only that this single-detection, hand-designed PHMM formulation approaches but does not beat a tuning-free NN time-weighted baseline. The AIS--RF distance gap difference persists under the log-odds score. One hypothesis is that the hand-designed $f_\delta$ does not match the generator's noise density and under-rewards the spatially closer candidate.

Per candidate, the PHMM Forward recursion costs $O(nk^2)$ with $k=5$ states and a single RF detection, and the log-odds score needs two passes (full and null model). NN time-weighted costs $O(n)$ with no state recursion.

\subsection{Threats to Validity}
\label{limitations}
Reliance on synthetic RF data limits how far the findings generalize to real-world conditions. We reproduced UnseenLabs' revisit patterns and introduced a Gaussian position error calibrated to UnseenLabs' reported geolocation accuracy ($\sigma=800$\,m, Sec.~\ref{rf-generation}), but the generator uses one global revisit schedule, no missed detections or false alarms, and an isotropic Gaussian error. $f_\varepsilon$ and $f_\delta$ were designed with knowledge of the generator, which may overstate performance on real noise.

Several design choices weaken the evidence. The log-odds score was introduced after inspecting validation-split results. It was not tuned on them, but the design choice was motivated by them. $\kappa$ was set informally, and the Forward baseline partly depends on it (Sec.~\ref{sec:phmm}). The bias tests use only misclassified detections, the log-odds score has fewer errors than corrected Forward and thus less power, and Experiments 1--3 use Welch's test, which ignores the pairing by detection.

A third limitation arises from the preprocessing of AIS data. Because trajectories were segmented at gaps longer than 551 seconds, the analysis is restricted to continuous tracks, which excludes vessels that go dark mid-voyage except near the start or end of the resulting segment. The dark-vessel test introduced in this study covers a related but different scenario, a vessel that has no AIS trajectory among the candidates, and does not address this limitation directly. Finally, the transition functions were hand-designed without input from maritime domain experts and are based solely on cargo vessels, so the model cannot be directly applied to other vessel types without retuning.

\section{Conclusion and Future Work}

To answer the question of how AIS and RF data can be matched to enhance ship identification, a Pair Hidden Markov Model (PHMM) was applied to align single RF detections with AIS trajectories, and compared with nearest-neighbor baselines in matching and in detecting dark vessels. On synthetic RF detections generated from NOAA AIS data, the length-corrected Forward score reached a precision of $0.7776$ on detections with multiple candidate AIS trajectories and a dark-vessel AUC of $0.637$. The correction reduced but did not remove the length bias. With a single RF detection, the corrected Forward score is largely a trajectory prior. A log-odds score, which compares the full model with a null model without match transitions, raised precision to $0.8580$ and AUC to $0.852$. A time-weighted nearest-neighbor baseline that requires no fitting remained ahead (precision $0.9013$, AUC $0.904$). This single-detection, hand-designed PHMM formulation therefore approaches but does not beat the baseline. A bias towards candidates with a larger AIS--RF distance remains under the length-corrected log-odds score ($n^{\alpha}$ variant). 

The identified limitations suggest directions for future research on the PHMM. Calibrating $f_\delta$ to the noise model should come first, since the persistent AIS--RF distance bias may stem from a mismatch between the two. Although the length correction improved alignment performance, substantial differences remain between short and much longer trajectories. The other biases of the Forward score, such as the preference for stationary trajectories, should be reduced as well. Applying weighting factors to adjust transition weights or improving the preselection strategy could reduce them. Recognizing that no true match exists deserves particular attention, since it is central to detecting the dark or AIS-manipulated vessels that motivate the use of RF data. Finally, additional AIS features not considered in this study should be incorporated into the model. For instance, vessel heading and status could be integrated into the transition weight functions or preselection approach.

\section*{Acknowledgment}
The MARIT-D project has received co-funding from the Internal Security Fund - Police programme under grant agreement n° 101114216.
During the preparation of this work, the authors used Claude Sonnet~5.5 (Anthropic) to refactor existing code and to partially generate new code in the accompanying GitHub repository. The authors reviewed and edited the code and take full responsibility for the content of the publication.

\bibliographystyle{IEEEtran}
\bibliography{main}

\end{document}

%% file: frames.tex
\newlength{\FramedXMargin}
\newlength{\FramedYMargin}
\newlength{\FramedSkipAbove}
\newlength{\FramedSkipBelow}

\newsavebox{\FramedBox}

\newenvironment{Framed}{%
  \begin{lrbox}{\FramedBox}
  \begin{minipage}{\dimexpr\linewidth-2\fboxrule-2\fboxsep} 
  \vspace{\FramedYMargin}%
  \centering%
  \begin{minipage}{\dimexpr\linewidth-2\FramedXMargin} 
}{%
  \end{minipage}%
  \vspace{\FramedYMargin}
  \end{minipage}%
  \end{lrbox}%
  \vspace{\FramedSkipAbove}
  \noindent\fbox{\usebox{\FramedBox}}
  \vspace{\FramedSkipBelow}
}